\documentclass[10pt,twocolumn,letterpaper]{article}

\usepackage{cvpr}
\usepackage{times}
\usepackage{epsfig}
\usepackage{graphicx}
\usepackage{amsmath}
\usepackage{amssymb}
\usepackage{subcaption}
\usepackage{etoolbox}

\usepackage[pagebackref=true,breaklinks=true,letterpaper=true,colorlinks,bookmarks=false]{hyperref}

\newcommand{\MyMapTemplatePrefixc}[4]{\expandafter#1\csname#3#4\endcsname{#2{#4}}}
\forcsvlist{\MyMapTemplatePrefixc {\def} {\mathcal}{c}} {A,B,C,D,E,F,G,H,I,J,K,L,M,N,O,P,Q,R,S,T,U,V,W,X,Y,Z}

\newcommand{\MyMapTemplatePrefixtb}[5]{\expandafter#1\csname#4#5\endcsname{#2{#3{#5}}}}
\forcsvlist{\MyMapTemplatePrefixtb {\def} {\tilde}{\mathbf}{t}} {A,B,C,D,E,F,G,H,I,J,K,L,M,N,O,P,Q,R,S,T,U,V,W,X,Y,Z}
\forcsvlist{\MyMapTemplatePrefixtb {\def} {\tilde}{\mathbf}{t}} {0,1,a,b,c,d,e,f,g,h,i,j,k,l,m,n,o,p,q,r,s,u,v,w,x,y,z}

\newcommand{\MyMapTemplateNoPrefix}[3]{\expandafter#1\csname#3\endcsname{#2{#3}}}
\forcsvlist{\MyMapTemplateNoPrefix {\def} {\mathbf} } {0,1,a,b,c,d,e, f, g, h, i, j, k, l, m, n, o, p, q, r, u, v, w, x, y, z}
\forcsvlist{\MyMapTemplateNoPrefix {\def} {\mathbf} } {A,B,C,D,E,F,G,H,I,J,K,L,M,N,O,P,Q,R,S,T,U,V,W,X,Y,Z}

\def\E{\mathbb{E}}

\cvprfinalcopy

\ifcvprfinal\fi
\begin{document}

\title{DLOW: Domain Flow for Adaptation and Generalization}

\author{
Rui Gong\textsuperscript{1}\space\space\space\space Wen Li\textsuperscript{1}\space\space\space\space Yuhua Chen\textsuperscript{1}\space\space\space\space Luc Van Gool\textsuperscript{1,2}\\
\textsuperscript{1}Computer Vision Laboratory, ETH Zurich\space\space\space\space \textsuperscript{2}VISICS, ESAT/PSI, KU Leuven\\
{\tt\small gongr@student.ethz.ch $\{$liwen, yuhua.chen, vangool$\}$@vision.ee.ethz.ch}
}

\maketitle
\thispagestyle{empty}

\begin{abstract}
In this work, we present a domain flow generation(DLOW) model to bridge two different domains by generating a continuous sequence of intermediate domains flowing from one domain to the other. The benefits of our DLOW model are two-fold. First, it is able to transfer source images into different styles in the intermediate domains. The transferred images smoothly bridge the gap between source and target domains, thus easing the domain adaptation task. 
Second, when multiple target domains are provided for training, our DLOW model is also able to generate new styles of images that are unseen in the training data.  We implement our DLOW model based on CycleGAN. A domainness variable is introduced to guide the model to generate the desired intermediate domain images. 
In the inference phase, a flow of various styles of images can be obtained by varying the domainness variable. We demonstrate the effectiveness of our model for both cross-domain semantic segmentation and the style generalization tasks on benchmark datasets.  Our implementation is available at \url{https://github.com/ETHRuiGong/DLOW}.
\end{abstract}

\section{Introduction}
The domain shift problem is drawing increasing attention in recent years \cite{Hoffman_cycada2017, zhu2017unpaired, Tsai_adaptseg_2018, sankaranarayanan2017unsupervised, ghifary2015domain, StarGAN2018}. In particular, there are two tasks that are of interest in computer vision community. One is the  \emph{domain adaptation} problem, where the goal is to learn a model for a given task from a label-rich data domain (\ie, source domain) to perform well in a label-scarce data domain (\ie, target domain). The other one is the \emph{image translation} problem, where the goal is to transfer images in the source domain to mimic the image style in the target domain. 

\begin{figure}[t]
\includegraphics[width=\linewidth]{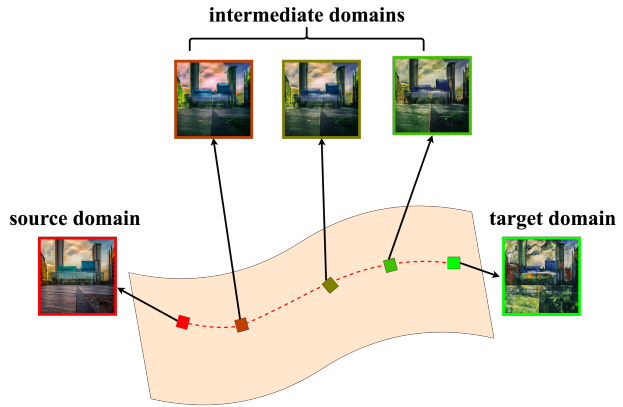}
\caption{Illustration of data flow generation. Traditional image translation methods directly map the image from the source domain to the target domain, while our DLOW model is able to produce a sequence of intermediate domains shifting 
from the source domain to the target domain.}
\label{DG_Transfer}
\end{figure}

Generally, most existing works focus on the target domain only. They aim to learn models that well fit the target data distribution, \eg, achieving good classification accuracy in the target domain, or transferring source images into the target style. In this work, we instead are interested in the intermediate domains between source and target domains. We present a new \emph{domain flow generation} (DLOW) model, which is able to translate images from the source domain into an arbitrary intermediate domain between source and target domains. As shown in Fig \ref{DG_Transfer}, by translating a source image along the domain flow from the source domain to the target domain, we obtain a sequence of images that naturally characterize the distribution shift from the source domain to the target domain. 

The benefits of our DLOW model are two-fold. First, those intermediate domains are helpful to bridge the distribution gap between two domains. By translating images into intermediate domains, those translated images can be used to ease the domain adaptation task. We show that the traditional domain adaptation methods can be boosted to achieve better performance in target domain with intermediate domain images. Moreover, the obtained models also exhibit good generalization ability on new datasets that are not seen in the training phase, benefiting from the diverse intermediate domain images. 

Second, our DLOW model can be used for style generalization. Traditional image-to-image translation works~\cite{zhu2017unpaired, pix2pix2017, pmlr-v70-kim17a, NIPS2017_6672} mainly focus on learning a deterministic one-to-one mapping that transfers a source image into the target style. In contrast, our DLOW model allows to translate a source image into an intermediate domain that is related to multiple target domains. For example, when performing the photo to painting transfer, instead of obtaining a Monet or Van Gogh style, our DLOW model could produce a mixed style of Van Gogh, Monet, etc. Such mixture can be customized in the inference phase by simply adjusting an input vector that encodes the relatedness to different domains. 

We implement our DLOW model based on CycleGAN~\cite{zhu2017unpaired}, which is one of the state-of-the-art unpaired image-to-image translation methods. We augment the CycleGAN to include an additional input of domainness variable. On one hand, the domainness variable is injected into the translation network using the conditional instance normalization layer to affect the style of output images. On the other hand, it is also used as weights on discriminators to balance the relatedness of the output images to different domains. For multiple target domains, the domainness variable is extended as a vector containing the relatedness to all target domains. Extensive results on benchmark datasets demonstrate the effectiveness of our proposed model for domain adaptation and style generalization. 

\section{Related Work}
\textbf{Image to Image Translation:}
Our work is related to the image-to-image translation works. The image-to-image translation task aims at translating the image from one domain into another domain. Inspired by the success of Generative Adversarial Networks(GANs)~\cite{goodfellow2014generative}, many works have been proposed to address the image-to-image translation based on GANs~\cite{pix2pix2017, wang2018pix2pixHD,zhu2017unpaired,NIPS2017_6672,lu2017conditional,he2017arbitrary,zhu2017toward,huang2018munit,almahairi2018augmented, StarGAN2018, DRIT, yi2017dualgan,lin2018conditional}. The early works~\cite{pix2pix2017, wang2018pix2pixHD} assume that  paired images between two domains are available, while the recent works such as CycleGAN~\cite{zhu2017unpaired}, DiscoGAN~\cite{pmlr-v70-kim17a} and UNIT~\cite{NIPS2017_6672} are able to train networks without using paired images. However, those works focus on learning deterministic image-to-image mappings. Once the model is learnt, a source image can only be transferred to a fixed target style. 

A few recent works~\cite{lu2017conditional,he2017arbitrary,zhu2017toward,huang2018munit,almahairi2018augmented, StarGAN2018, DRIT, yi2017dualgan, lin2018conditional, lample2017fader} concentrate on learning a unified model to translate images into multiple styles. These works can be divided into two categories according to the controllability of the target styles. The first category, such as \cite{huang2018munit, almahairi2018augmented}, realizes the multimodal translation by sampling different style codes which are encoded from the target style images. However, those works focus on modelling intra-domain diversity, while our DLOW model aims at characterizing the inter-domain diversity. Moreover, they cannot explicitly control the translated target style using the input codes. 

The second category, such as \cite{StarGAN2018, lample2017fader}, assigns the domain labels to different target domains and the domain labels are proven to be effective in controlling the translation direction. Among those, \cite{lample2017fader} shows that they could make interpolation between target domains by continuously shifting the different domain labels to change the extent of the contribution of different target domains. However, these methods only use the discrete binary domain labels in the training. Unlike the above work, the domainness variable proposed in this work is derived from the data distribution distance, and is used explicitly to regularize the style of output images during training. 

\textbf{Domain Adaptation and Generalization:}
Our work is also related to the domain adaptation and generalization works. Domain adaptation aims to utilize a labeled source domain to learn a model that performs well on an unlabeled target domain~\cite{ganin2015unsupervised, gopalan2011domain, fernando2013unsupervised, tzeng2017adversarial, jhuo2012robust, baktashmotlagh2013unsupervised, kodirov2015unsupervised, gong2012geodesic, chen2018domain, zhang2018collaborative, wulfmeier2018incremental}. Domain generalization is a similar problem, which aims to learn a model that could be generalized to an unseen target domain by using multiple labeled source domains~\cite{pmlr-v28-muandet13, ghifary2015domain, niu2015visual, motiian2017unified, niu2015multi,Li2018Domain, li2018deep,li2018domain_gene}. 

Our work is partially inspired by \cite{gopalan2011domain,gong2012geodesic,cui2014flowing}, which have shown that the intermediate domains between source and target domains are useful for addressing the domain adaptation problem. They represent each domain as a subspace or covariance matrix, and then connect them on the corresponding manifold to model intermediate domains. Different from those works, we model the intermediate domains by directly translating images on pixel level. This allows us to easily improve the existing deep domain adaptation models by using the translated images as training data. Moreover, our model can also be applied to image-level domain generalization by generating mixed-style images. 

Recently, there is an increasing interest to apply domain adaptation techniques for semantic segmentation from synthetic data to the real scenario \cite{hoffman2016fcns, Hoffman_cycada2017, chen2018road, zou2018unsupervised, luo2018taking, huang2018domain, dundar2018domain, pan2018IBN-Net, saleh2018effective, sankaranarayanan2018learning, hong2018conditional, peng2018visda, zhang2018fully, Tsai_adaptseg_2018, murez2017image, saito2017maximum, sankaranarayanan2017unsupervised, zhu2018penalizing, chen2018learning}. Most of those works conduct the domain adaptation by adversarial training on the feature level with different priors. The recent Cycada \cite{Hoffman_cycada2017} also shows that it is beneficial to perform pixel-level domain adaptation firstly by transferring source image into the target style based on the image-to-image translation methods like CycleGAN \cite{zhu2017unpaired}. However, those methods address domain shift by adapting to only the target domain. In contrast, we aim to perform pixel-level adaptation by transferring source images to a flow of intermediate domains. Moreover, our model can also be used to further improve the existing feature-level adaptation methods. 

\section{Domain Flow Generation}

\subsection{Problem Statement}
In the domain shift problem, we are given a source domain $\cS$ and a target domain $\cT$ containing samples from two different distributions $P_S$ and $P_T$, respectively. Denoting a source sample as $\x^s \in \cS$ and a target sample as $\x^t \in \cT$, we have  $\x^s \sim P_S$,  $\x^t \sim P_T$, and  $P_S \neq P_T$.

Such distribution mismatch usually leads to a significant performance drop when applying the model trained on $\cS$ to $\cT$. Many works have been proposed to address the domain shift for different vision applications. A group of recent works aim to reduce the distribution difference on the feature level by learning domain-invariant features~\cite{ganin2015unsupervised, gopalan2011domain, kodirov2015unsupervised, gong2012geodesic}, while others work on the image level to transfer source images to mimic the target domain style~\cite{zhu2017unpaired,NIPS2017_6672,zhu2017toward,huang2018munit,almahairi2018augmented, StarGAN2018}. 

In this work, we also propose to address the domain shift problem on image level. However, different from existing works that focus on transferring source images into only the target domain, we instead transfer them into all intermediate domains that connect source and target domains. This is partially motivated by the previous works \cite{gopalan2011domain,gong2012geodesic,cui2014flowing}, which have shown that the intermediate domains between source and target domains are useful for addressing the domain adaptation problem.

In the follows, we first briefly review the conventional image-to-image translation model CycleGAN. Then, we formulate the intermediate domain adaptation problem based on the data distribution distance. Next, we present our DLOW model based on the CycleGAN model. We then show the benefits of our DLOW model with two applications: 1) improve existing domain adaptation models with the images generated from DLOW model, and 2) transfer images into arbitrarily mixed styles when there are multiple target domains. 

\subsection{The CycleGAN Model}
We build our model based on the state-of-the-art CycleGAN model~\cite{zhu2017unpaired} which is proposed for unpaired image-to-image translation. Formally, the CycleGAN model learns two mappings between $\cS$ and $\cT$, \ie, $G_{ST}: \cS\rightarrow\cT$ which transfers the images in $\cS$ into the style of $\cT$, and $G_{TS}: \cT\rightarrow\cS$ which acts in the inverse direction. We take the $\cS\rightarrow\cT$ direction as an example to explain CycleGAN.

To transfer source images into the target style and also preserve the semantics, the CycleGAN employs an adversarial training module and a reconstruction module, respectively. In particular, the adversarial training module is used to align the image distributions for two domains, such that the style of mapped images matches the target domain. Let us denote the discriminator as $D_T$, which attempts to distinguish the translated images and the target images. Then the objective function of the adversarial training module can be written as,
\begin{eqnarray}
\label{eqn:cyclegan_adv}
\min_{G_{ST} }\max_{D_T} \!\!\!\!&& \!\!\!\!\E_{\x^t\sim{P_T}}\left[\log(D_{T}(\x^t))\right]\\
\!\!\!\!&+& \!\!\!\!\E_{\x^s\sim{P_S}}\left[\log(1-D_{T}(G_{ST}(\x^s)))\right].\nonumber
\end{eqnarray}
Moreover, the reconstruction module is to ensure the mapped image $G_{ST}(\x^s)$ to preserve the semantic content of the original image $\x^s$. This is achieved by enforcing a cycle consistency loss such that $G_{ST}(\x^s)$ is able to recover $\x^s$ when being mapped back to the source style, \ie, 
\begin{eqnarray}
\min_{G_{ST}} \quad \E_{\x^s\sim{P_S}}\left[\|G_{TS}(G_{ST}(\x^s))-\x^s\|_{1}\right].
\end{eqnarray}
Similar modules are applied to the $\cT\rightarrow\cS$ direction. By jointly optimizing all modules, CycleGAN model is able to transfer source images into the target style and v.v.
\begin{figure}
\centering
\includegraphics[width=0.7\linewidth]{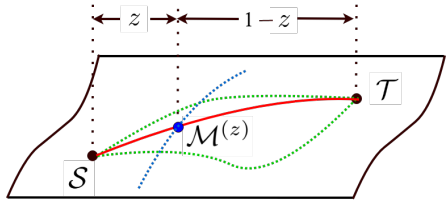}
\caption{Illustration of domain flow. Many possible paths (the green dash lines) connect source and target domains, while the domain flow is the shortest one (the red line). There are multiple domains (the blue dash line) keeping the expected relative distances to source and target domains. An intermediate domain (the blue dot) is the point at the domain flow that keeps the right distances to two domains.}
\label{optimized_path}
\vspace{-10pt}
\end{figure}

\begin{figure*}[!ht]
\centering
  \centering
  \includegraphics[height=0.23\paperheight]{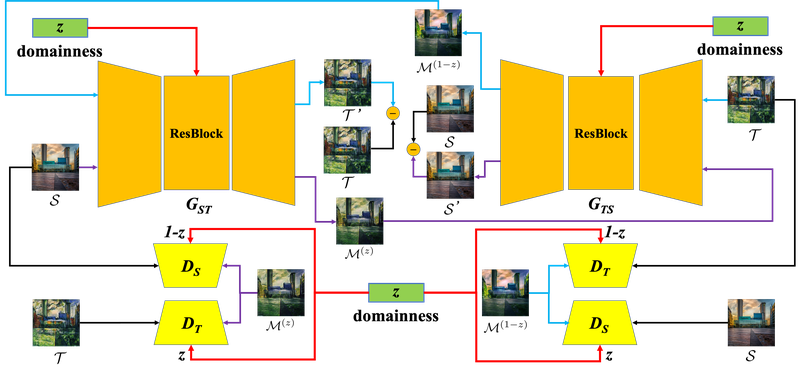}
  \label{fig_domaininter}
\setlength{\abovecaptionskip}{0pt} 
\caption{The overview of our DLOW model: the generator takes domainness $z$ as additional input to control the image translation and to reconstruct the source image; The domainness $z$ is also used to weight the two discriminators.}
\vspace{-10pt}
\label{fig_domaininter:fig}
\end{figure*}

\subsection{Modeling Intermediate Domains}\label{sec:intermediate}
Intermediate domains have been shown to be helpful for domain adaptation  \cite{gopalan2011domain,gong2012geodesic,cui2014flowing}, where they model intermediate domains as a geodesic path on Grassmannian or Riemannian manifold. Inspired by those works, we also characterize the domain shift using intermediate domains that connect the source and target domains. Diffrent from those works, we directly operate at the image level, \ie, translating source images into different styles corresponding to intermediate domains. In this way, our method can be easily integrated with deep learning techniques for enhancing the cross-domain generalization ability of models. 

In particular, let us denote an intermediate domain as $\cM^{(z)}$, where $z \in [0, 1]$ is a continous variable which models the relatedness to source and target domains. We refer to $z$ as the domainness of intermediate domain. When $z = 0$, the intermediate domain $\cM^{(z)}$ is identical to the source domain $\cS$; and when $z = 1$, it is identical to the target domain $\cT$. By varying $z$ in the range of $[0, 1]$, we thus obtain a sequence of intermediate domains that flow from $\cS$ to $\cT$.

There are many possible paths to connect the source and target domains. As shown in Fig \ref{optimized_path}, assuming there is a manifold of domains, where a domain with given data distribution can be seen as a point residing at the manifold. We expect the domain flow $\cM^{(z)}$ to be the shortest geodesic path connecting $\cS$ and $\cT$. Moreover, given any $z$, the distance from $\cS$ to $\cM^{(z)}$ should also be proportional to the distance between $\cS$ to $\cT$ by the value of $z$. Denoting the data distribution of $\cM^{(z)}$ as $P_M^{(z)}$, we expect that,
\begin{equation}
    \begin{aligned}
    \frac{dist\left(P_S, P_M^{(z)}\right)}{dist\left(P_T, P_M^{(z)}\right)}=\frac{z}{1-z},
    \end{aligned}
    \label{equation_domainness}
\end{equation}
where $dist(\cdot, \cdot)$ is a valid distance measurement over two distributions. Thus, generating an intermediate domain $\cM^{(z)}$ for a given $z$ becomes finding the point satisfying Eq. (\ref{equation_domainness}) that is closet to $\cS$ and $\cT$, which leads to minimize the following loss,
\begin{eqnarray}
\cL = (1-z) \cdot dist\left(P_S, P_M^{(z)}\right) + z\cdot dist\left(P_T, P_M^{(z)}\right).
\label{eqn:intermediate_domain}
\end{eqnarray}
As shown in \cite{arjovsky2017wasserstein}, many types of distance have been exploited for image generation and image translation. The adversarial loss in Eq.~(\ref{eqn:cyclegan_adv}) can be seen as a lower bound of the Jessen-Shannon divergence. We also use it to measure distribution distance in this work.

\subsection{The DLOW Model}
We now present our DLOW model to generate intermediate domains. Given a source image $\x^s \sim P_s$, and a domainness variable $z \in [0, 1]$, the task is to transfer $\x^s$ into the intermediate domain $\cM^{(z)}$ with the distribution $P_M^{(z)}$ that minimizes the objective in Eq.~(\ref{eqn:intermediate_domain}). We take the $\cS\rightarrow\cT$ direction as an example, and the other direction can be similarly applied. 

In our DLOW model, the generator $G_{ST}$ no longer aims to directly transfer $\x^s$ to the target domain $\cT$, but to move $\x^s$ towards it. The interval of such moving is controlled by the domainness variable $z$. Let us denote $\cZ = [0, 1]$ as the domain of $z$, then the generator in our DLOW model can be represented as $G_{ST}(\x^s, z): \cS \times \cZ \rightarrow\cM^{(z)}$ where the input is a joint space of $\cS$ and $\cZ$. 

\textbf{Adversarial Loss:}  As discussed in Section~\ref{sec:intermediate}, We deploy the adversarial loss as the distribution distance measurement to control the relatedness of an intermediate domain to the source and target domains. Specifically, we introduce two discriminators, $D_{S}(\x)$ to distinguish $\cM^{(z)}$  and $\cS$, and $D_{T}(\x)$ to distinguish $\cM^{(z)}$  and $\cT$, respectively. Then, the adversarial losses between $\cM^{(z)}$ and $\cS$ and $\cT$ can be written respectively as,
\begin{eqnarray}
\!\!\!\!\cL_{adv} (\!\!\!\!&G_{ST}&\!\!\!\!, D_S)=\E_{\x^s\sim{P_S}}\left[\log(D_{S}(\x^s))\right] \\ 
\!\!\!\!&+&\!\!\!\!\E_{\x^s\sim{P_S}}\left[\log(1-D_{S}(G_{ST}(\x^s, z)))\right] \nonumber \\
\!\!\!\!\cL_{adv} (\!\!\!\!&G_{ST}&\!\!\!\!, D_T)=\E_{\x^t\sim{P_T}}\left[\log(D_{T}(\x^t))\right]  \\
\!\!\!\!&+&\!\!\!\!\E_{\x^s\sim{P_S}}\left[\log(1-D_{T}(G_{ST}(\x^s, z)))\right]. \nonumber
\end{eqnarray}

By using the above losses to model $dist(P_S, P_M^{(z)})$ and $dist(P_T, P_M^{(z)})$ in Eq.~(\ref{eqn:intermediate_domain}), we derive the following loss,
\begin{eqnarray}
\cL_{adv}  = (1-z) \cL_{adv}(G_{ST}, D_S) + z \cL_{adv}(G_{ST}, D_T).
\end{eqnarray}

\textbf{Image Cycle Consistency Loss:} Similarly as in CylceGAN, we also apply a cycle consistency loss to ensure the semantic content is well-preserved in the translated image. Let us denote the generator on the other direction as $G_{TS}(\x^t, z): \cT \times \cZ \rightarrow\cM^{(1-z)}$, which transfers a sample $\x^t$ from the target domain towards the source domain by a interval of $z$. Since $G_{TS}$ acts in an inverse direction to $G_{ST}$, we can use it to recover $\x^s$ from the translated version $G_{ST}(\x^s, z)$, which gives the following loss,
\begin{equation}
\begin{aligned}
L_{cyc} = &\E_{\x^s\sim{P_s}}\left[\|G_{TS}(G_{ST}(\x^s,z),z)-\x^s\|_{1}\right].
\end{aligned}
\end{equation}

\textbf{Full Objective:} We integrate the losses defined above, then the full objective can be defined as, 
\begin{equation}
\begin{aligned}
 &\cL=\cL_{adv} + \lambda_1\cL_{cyc},
\end{aligned}
\end{equation}
where $\lambda_{1}$ is a hyper-parameter used to balance the two losses in the training process. 

Similar loss can be defined for the other direction $\cT\rightarrow\cS$. Due to the usage of adversarial loss $\cL_{adv}$, the training is performed in an alternating manner. We first minimize the full objective with regard to the generators, and then maximize it with regard to the discriminators. 

\textbf{Implementation: } We illustrate the network structure of of our DLOW model in Fig \ref{fig_domaininter:fig}. First, the domainness variable $z$ is taken as the input of the generator $G_{ST}$. This is implemented with the Conditional Instance Normalization (CN) layer~\cite{almahairi2018augmented,huang2017adain}. We first use one deconvolution layer to map the domainness variable $z$ to the vector with dimension $(1,16,1,1)$, and then use this vector as the input for the CN layer. Moreover, the domainness variable also plays the role of weighting discriminators to balance the relatedness of the generated images to different domains. It is also used as input in the image cycle consistency module. During the training phase, we randomly generate the domainess parameter $z$ for each input image. As inspired by \cite{zhang2018mixup},  we force the domainness variable $z$ to obey the beta distribution, i.e. $f(z,\alpha, \beta)=\frac{1}{B(\alpha,\beta)}z^{\alpha-1}(1-z)^{\beta-1}$, where $\beta$ is fixed as $1$, and $\alpha$ is a function of the training step $\alpha=e^{\frac{t-0.5T}{0.25T}}$ with $t$ being the current iteration and $T$ being the total number of iterations. In this way, $z$ tends to be sampled more likely as small values at the beginning, and gradually shift to larger values at the end, which gives slightly more stable training than uniform sampling.

\subsection{Boosting Domain Adaptation Models}\label{sec:boostadaptation}
With the DLOW model, we are able to translate each source image $\x^s$ into an arbitrary intermediate domain $\cM^{(z)}$. Let us denote the source dataset as $\cS = \{(\x^s_i, y_i)|_{i=1}^n\}$ where $y_i$ is the label of $\x^s_i$. By feeding each of the image $\x^s_i$ combined with $z_{i}$ randomly sampled from the uniform distribution $\cU(0,1)$, we then obtain a translated dataset $\tilde{\cS} = \{(\tx^s_i, y_i)|_{i=1}^n\}$ where $\tx^s_i = G_{ST}(\x^s_i, z_i)$ is the translated version of $\x^s_i$. The images in $\tilde{\cS}$ spread along the domain flow from source to target domain, and therefore become much more diverse. Using $\tilde{\cS}$ as the training data is helpful to learn domain-invariant models for computer vision tasks. In Section~\ref{sec:exp_da}, we demonstrate that model trained on $\tilde{\cS}$ achieves good performance for the cross-domain semantic segmentation problem. 

Moreover, the translated dataset $\tilde{\cS}$ can also be used to boost the existing adversarial training based domain adaptation approaches. 
Images in $\tilde{\cS}$ fill the gap between the source and target domains, and thus ease the domain adaptation task. 
Taking semantic segmentation as an example, a typical way is to append a discriminator to the segmentation model, which is used to distinguish the source and target samples. Using the adversarial training strategy to optimize the discriminator and the segmentation model, the segmentation model is trained to be more domain-invariant. 

\begin{figure}
\includegraphics[width=\linewidth]{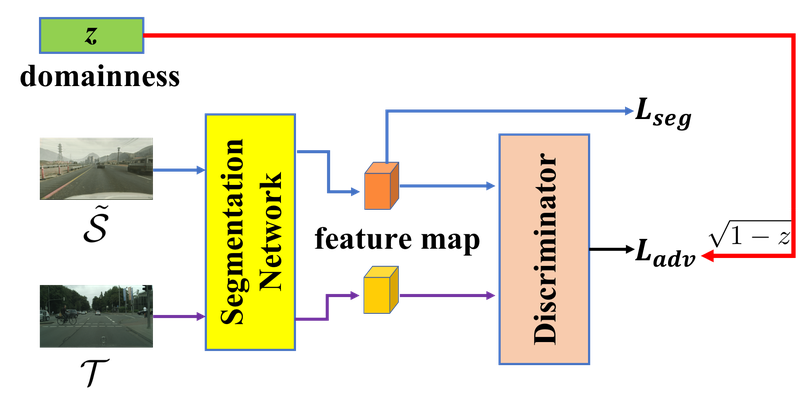}
\caption{Illustration of boosting domain adaptation model for corss-domain semantic segmentation with DLOW model. Intermediate domain images are used as source dataset, and the adversarial loss is weighted by domainness.}
\label{domainnessadaptseg}
\vspace{-10pt}
\end{figure}

As shown in Fig~\ref{domainnessadaptseg}, we replace the source dataset ${\cS}$ with the translated version $\tilde{\cS}$, and apply a weight $\sqrt{1-z_i}$ to the adversarial loss. The motivation is as follows, for each sample $\tx^s_i$, if the domainness $z_i$ is higher, it is closer to the target domain, then the weight of adversarial loss can be reduced. Otherwise, we should enhance the loss weight. 

\subsection{Style Generalization}\label{sec:stylegeneralization}
Most existing image-to-image translation works learn a deterministic mapping between two domains. After the model is learnt, source images can only be translated to a fixed style. In contrast, our DLOW model takes an random $z$ to translate images into various styles. When multiple target domains are provided, it is also able to transfer the source image into a mixture of different target styles. In other words, we are able to generalize to an unseen intermediate domain that is related to existing domains.

In particular, suppose we have $K$ target domains, denoted as $\cT_1, \ldots, \cT_K$. Accordingly, the domainness variable $z$ is expanded as a $K$-dim vector $\z = [z_1, \ldots, z_K]'$ with $\sum_{k=1}^Kz_k = 1$. Each elelment $z_k$ represents the relatedness to the $k$-th target domain. To map an image from the source domain to the intermediate domain defined by $\z$, we need to optimize the following objective,
\begin{eqnarray}
\cL = \sum_{k=1}^K z_{k} \cdot dist(P_M, P_{T_k}), \quad\mbox{s.t.}\quad \sum_{1}^K z_k = 1
\end{eqnarray}
where $P_M$ is the distribution of the intermediate domain, $P_{T_K}$ is the distribution of $T_k$. The network structure can be easily adjusted from our DLOW model to optimize the above objective. We leave the details in the Supplementary due to the space limitation. 

\begin{figure*}[t]
\centering
\begin{subfigure}[c]{0.19\textwidth}
  \centering
  \includegraphics[width=\linewidth]{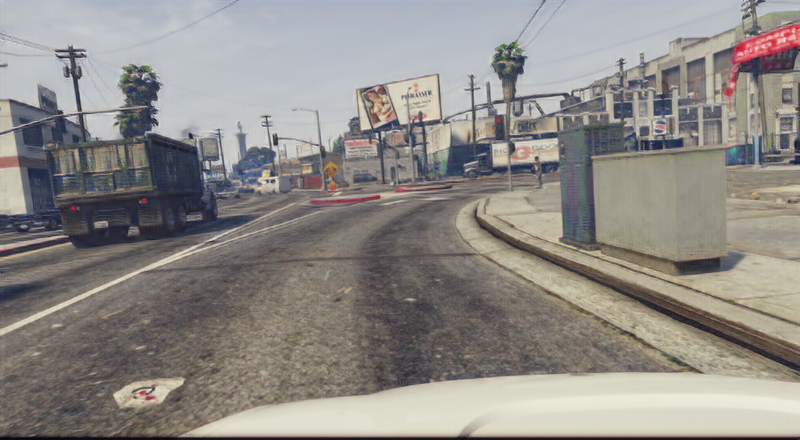}
  \caption{$z=0$}
  \label{fig_inter:sfig1}
\end{subfigure}
\begin{subfigure}[c]{0.19\textwidth}
  \centering
  \includegraphics[width=\linewidth]{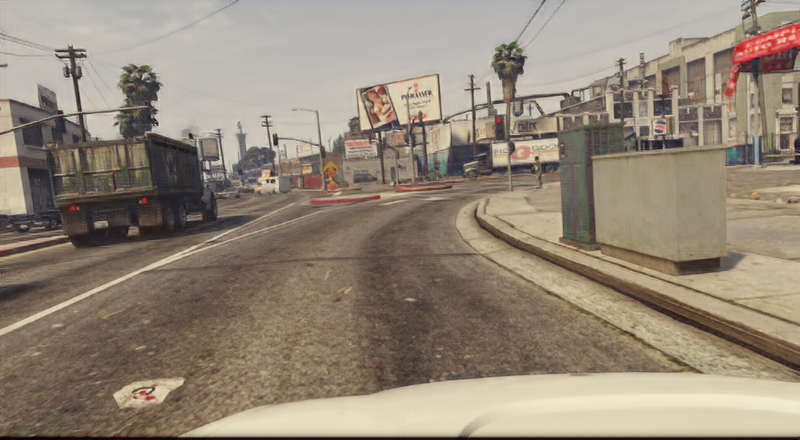}
  \caption{$z=0.3$}
  \label{fig_inter:sfig2}
\end{subfigure}
\begin{subfigure}[c]{0.19\textwidth}
  \centering
  \includegraphics[width=\linewidth]{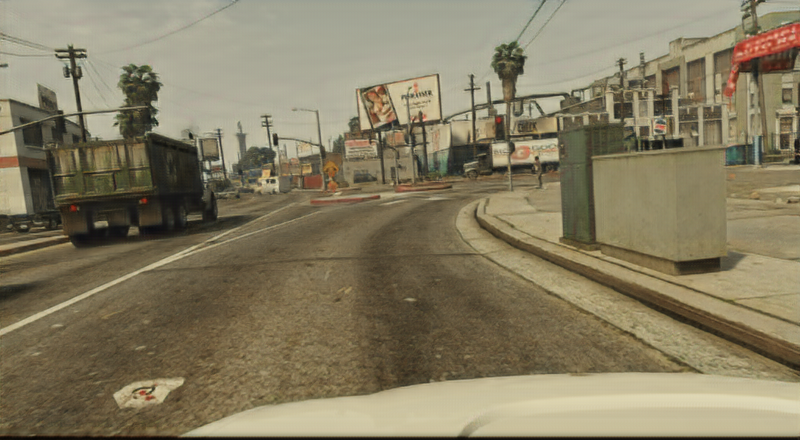}
  \caption{$z=0.6$}
  \label{fig_inter:sfig3}
\end{subfigure}
\begin{subfigure}[c]{0.19\textwidth}
  \centering
  \includegraphics[width=\linewidth]{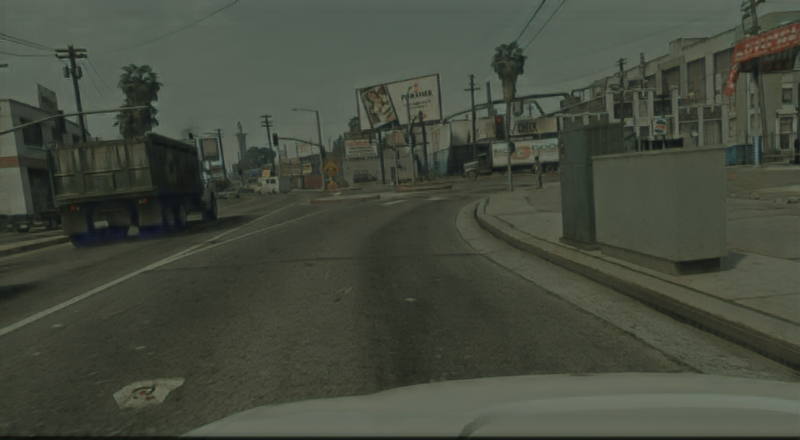}
  \caption{$z=0.8$}
  \label{fig_inter:sfig4}
\end{subfigure}
\begin{subfigure}[c]{0.19\textwidth}
  \centering
  \includegraphics[width=\linewidth]{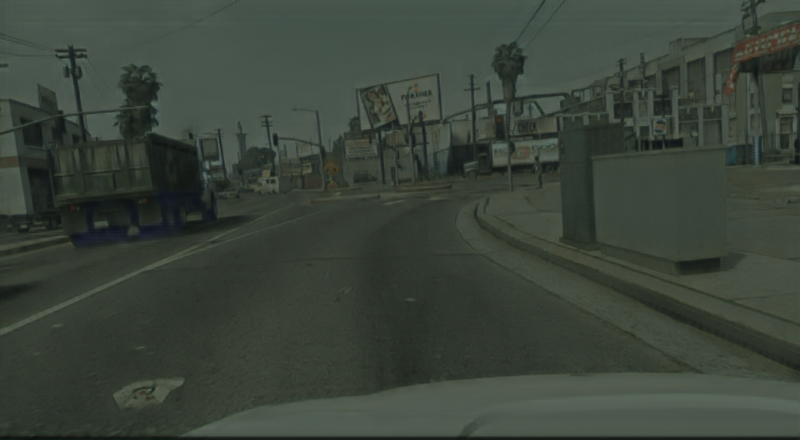}
  \caption{$z=1$}
  \label{fig_inter:sfig5}
\end{subfigure}
\caption{Examples of intermediate domain images from GTA5 to Cityscapes. As the domainness variable increases from 0 to 1, the styles of the translated images shift from the synthetic GTA5 style to the realistic Cityscapes style gradually.}
\label{fig_inter:fig}
\end{figure*}

\begin{table*}[h]
\setlength{\tabcolsep}{3pt}
\centering
 \resizebox{\textwidth}{20mm}
 {
 \begin{tabular}{c|ccccccccccccccccccc|c}
  \hline
  \multicolumn{21}{c}{GTA5 $\rightarrow$ Cityscapes}\\
  \hline
  Method&\rotatebox{90}{road}&\rotatebox{90}{sidewalk}&\rotatebox{90}{building}&\rotatebox{90}{wall}&\rotatebox{90}{fence}&\rotatebox{90}{pole}&\rotatebox{90}{traffic light}&\rotatebox{90}{traffic sign}&\rotatebox{90}{vegetation}&\rotatebox{90}{terrian}&\rotatebox{90}{sky}&\rotatebox{90}{person}&\rotatebox{90}{rider}&\rotatebox{90}{car}&\rotatebox{90}{truck}&\rotatebox{90}{bus}&\rotatebox{90}{train}&\rotatebox{90}{motorbike}&\rotatebox{90}{bicycle}&mIoU\\
  \hline
  NonAdapt\cite{Tsai_adaptseg_2018}&75.8&16.8&77.2&12.5&\textbf{21.0}&25.5&\textbf{30.1}&20.1&81.3&24.6&70.3&53.8&\textbf{26.4}&49.9&17.2&25.9&\textbf{6.5}&\textbf{25.3}&\textbf{36.0}&36.6\\
  CycleGAN\cite{Hoffman_cycada2017}&81.7&27.0&\textbf{81.7}&\textbf{30.3}&12.2&28.2&25.5&27.4&82.2&\textbf{27.0}&77.0&\textbf{55.9}&20.5&\textbf{82.8}&30.8&38.4&0.0&18.8&32.3&41.0\\
  \hline
  DLOW($z=1$)&\textbf{88.5}&\textbf{33.7}&80.7&26.9&15.7&27.3&27.7&\textbf{28.3}&80.9&26.6&74.1&52.6&25.1&76.8&30.5&27.2&0.0&15.7&\textbf{36.0}&40.7\\
  DLOW &87.1&33.5&80.5&24.5&13.2&\textbf{29.8}&29.5&26.6&\textbf{82.6}&26.7&\textbf{81.8}&\textbf{55.9}&25.3&78.0&\textbf{33.5}&\textbf{38.7}&0.0&22.9&34.5&\textbf{42.3}\\
  \hline
 \end{tabular}
}
\caption{\label{table_pixel} Results of semantic segmentation on the CityScapes dataset based on DeepLab-v2 model with ResNet-101 backbone using the images translated with different models. The results are reported on mIoU over 19 categories. The best result is denoted in bold.}
\vspace{-10pt}
\end{table*}
\begin{table}
\setlength{\tabcolsep}{3pt}
\centering
 \begin{tabular}{c|cccc}
  \hline
  &Cityscapes&KITTI&WildDash&BDD100K\\
  \hline
  Original~\cite{Tsai_adaptseg_2018}&42.4&30.7&18.9&37.0\\
  DLOW&\textbf{44.8}&\textbf{36.6}&\textbf{24.9}&\textbf{39.1}\\
  \hline
 \end{tabular}
 \caption{\label{table_generalization} Comparison of the performance of AdaptSegNet~\cite{Tsai_adaptseg_2018} when using original source images and intermediate domain images translated with our DLOW model for semantic segmention under domain adaptation (1st column) and domain generalization (2nd to 4th columns) scenarios. The results are reported on mIoU over 19 categories. The best result is denoted in bold.}
 \vspace{-10pt}
\end{table}

\section{Experiments}
In this section, we demonstrate the benefits of our DLOW model with two tasks. In the first task, we address the domain adaptation problem, and train our DLOW model to generate the intermediate domain samples to boost the domain adaptation performance. In the second task, we consider the style generalization problem, and train our DLOW model to transfer images into new styles that are unseen in the training data.

\subsection{Domain Adaptation and Generalization}\label{sec:exp_da}
\subsubsection{Experiments Setup}
For the domain adaptation problem, we follow~\cite{hoffman2016fcns, Hoffman_cycada2017, chen2018road, zou2018unsupervised} to conduct experiments on the urban scene semantic segmentation by learning from synthetic data to real scenario. The GTA5 dataset~\cite{Richter_2016_ECCV} is used as the source domain while the Cityscapes dataset~\cite{Cordts2016Cityscapes} as the target domain. Moreover, we also evaluate the generalization ability of learnt segmentation models to unseen domains, for which we take the KITTI~\cite{Geiger2012CVPR}, WildDash~\cite{Zendel_2018_ECCV} and BDD100K~\cite{yu2018bdd100k} datasets as additional unseen datasets for evaluation. We also conduct experiments using the SYNTHIA dataset~\cite{Ros_2016_CVPR} as the source domain, and provide the results in Supplementary.

\textbf{Cityscapes} is a dataset consisting of urban scene images taken from some European cities. We use the $2,993$ training images without annotation as unlabeled target samples in training phase, and 500 validation images with annotation for evaluation, which are densely labelled with 19 classes.

\textbf{GTA5} is a dataset consisting of $24,966$ densely labelled synthetic frames generated from the computer game whose scenes are based on the city of Los Angeles. The annotations of the images are compatible with the Cityscaps.

\textbf{KITTI} is a dataset consisting of images taken from mid-size city of Karlsruhe. We use 200 validation images densely labeled and compatible with Cityscapes.

\textbf{WildDash} is a dataset covers images from different sources, different environments(place, weather, time and so on) and different camera characteristics. We use 70 labeled and Cityscapes annotation compatible validation images.

\textbf{BDD100K} is a driving dataset covering diverse images taken from US whose label maps are with training indices specified in Cityscapes. We use $1,000$ densely labeled images for validation in our experiment.

In this task, we first train our proposed DLOW model using the GTA5 dataset as the source domain, and Cityscapes as the target domain. Then, we generate a translated GTA5 dataset with the learnt DLOW model. Each source image is fed into DLOW with a random domainness variable $z$. The new translated GTA5 dataset contains exactly the same number of images as the original one, but the styles of images randomly drift from the synthetic style to the real style. We then use the translated GTA dataset as the new source domain to train segmentation models.

We implement our model based on Augmented CycleGAN \cite{almahairi2018augmented} and CyCADA~\cite{Hoffman_cycada2017}. Following their setup, all images are resized to have width $1024$ while keeping the aspect ratio and the crop size is set as $400\times 400$. When training the DLOW model, the image cycle consistency loss weight is set as 10. The learning rate is fixed as 0.0002.
For the segmentation network, we use the AdaptSegNet~\cite{Tsai_adaptseg_2018} model, which is based on DeepLab-v2~\cite{chen2018deeplab} with ResNnet-101~\cite{he2016deep} as the backbone network.  The training images are resized to $1280\times 720$. We follow the exact the same training policy as in the AdaptSegNet.

\subsubsection{Experimental Results}
\textbf{Intermediate Domain Images:} To verify the ability of our DLOW model to generate intermediate domain images, in the inference phase, we fix the input source image, and vary the domainness variable from 0 to 1. A few examples are shown in Fig \ref{fig_inter:fig}. It can be observed that the styles of translated images gradually shift from the synthetic style of GTA5 to the real style of Cityscapes, which demonstrates the DLOW model is capable of modeling the domain flow to bridge the source and target domains as expected. Enlarged images and more discussion are provided in Supplementary.

\begin{figure*}
  \centering
  \includegraphics[height=0.3\paperheight]{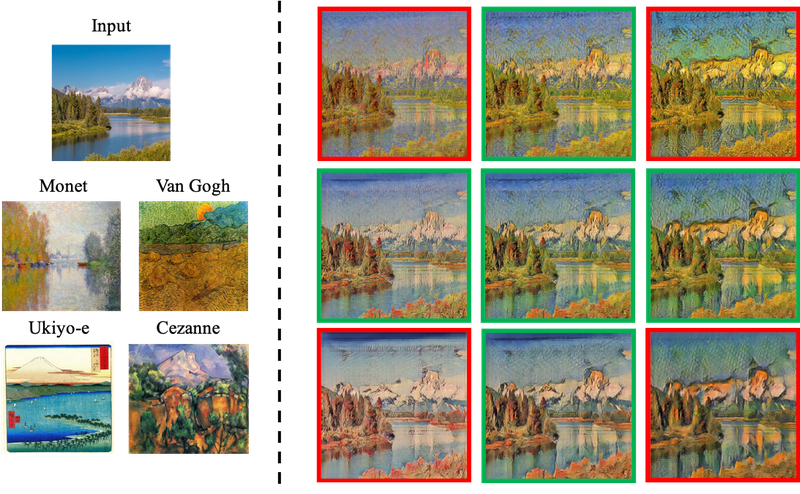}
  \caption{Examples of style generalization. Results with red rectangles at four corners are images translated into the four target domains, and those with green rectangles in between are images translated into intermediate domains. The results show that our DLOW model generalizes well across styles, and produces new images styles smoothly.}
  \label{fig_example}
  \vspace{-10pt}
\end{figure*}

\noindent \textbf{Cross-Domain Semantic Segmentation:} We further evaluate the usefulness of intermediate domain images in two settings. In the first setting, we compare with the CycleGAN model~\cite{zhu2017unpaired}, which is used in the CycADA approach~\cite{Hoffman_cycada2017} for performing pixel-level domain adaptation. The difference between CycleGAN and our DLOW model is that CycleGAN transfers source images to mimic only the target style, while our DLOW model transfers source images into random styles flowing from the source domain to the target domain. We first obtain a translated version of the GTA5 dataset with each model. Then, we respectively use the two transalated GTA5 datasets to train DeepLab-v2 models, which are evaluated on the Cityscapes dataset for semantic segmentation. We also include the ``NonAdapt" baseline which uses the original GTA5 images as training data, as well as a special case of our approach, ``DLOW($z=1$)", where we set $z = 1$ for all source images when making image translation using the learnt DLOW model. 

The results are shown in Table \ref{table_pixel}. We observe that all pixel-level adaptation methods outperform the ``NonAdapt" baseline, which verifies that image translation is helpful for training models for cross-domain semantic segmentation. Moreover, ``DLOW($z=1$)" is a special case of our model that directly translates source images into the target domain, which non-surprisingly gives comparable result as the CycADA-pixel method ($40.7\%$ v.s. $41.0\%$). By further using intermediate domain images, our DLOW model is able to improve the result from $40.7\%$ to $42.3\%$, which demonstrates that intermediate domain images are helpful for learning a more robust domain-invariant model. 

In the second setting, we further use intermediate domain images to improve the feature-level domain adpatation model. We conduct experiments based on the AdaptSegNet method~\cite{Tsai_adaptseg_2018}, which is open source and has reported the state-of-the-art result for GTA5$\rightarrow$CityScapes. It consists of multiple levels of adversarial training, and we augment each level with the loss weight discussed in Section~\ref{sec:boostadaptation}. The results are reported in Table~\ref{table_generalization}. The ``Original" method denotes the AdaptSegNet model that is trained using GTA5 as the source domain, for which the results are obtained using their released pretrained model. The ``DLOW" method is AdaptSegNet trained using translated dataset with our DLOW model. From the first column, we observe that the intermediate domain images are able to improve the AdaptSegNet model by $2.5\%$ from $42.3\%$ to $44.8\%$. More interestingly, we show that the AdaptSegNet model with DLOW translated images also exhibits excellent domain generalization ability when being applied to unseen domains, which achieves significantly better results than the original AdaptSegNet model on the KITTI, WildDash and BDD100K datasets as reported in the second to the fourth columns, respectively. This shows that intermediate domain images are useful to improve the model's cross-domain generalization ability.

\subsection{Style Generalization}
We conduct the style generalization experiment on the Photo to Artworks dataset\cite{zhu2017unpaired}, which consists of real photographs ($6,853$ images) and artworks from Monet($1,074$ images), Cezanne($584$ images), Van Gogh($401$ images) and Ukiyo-e($1,433$ images). We use the real photographs as the source domain, and the remaining as four target domains. 
As discussed in Section~\ref{sec:stylegeneralization}, The domainness variable in this experiment is expanded as a $4$-dim vector $[z_1, z_2, z_3, z_4]'$ meeting the condition $\sum_{i=1}^{4}z_i=1$. Also, $z_{1}, z_{2}, z_{3}$ and $z_{4}$ corresponds to Monet, Van Gogh, Ukiyo-e and Cezanne, respectively. Each element $z_i$ can be seen as how much each style contributes to the final mixture style. 
In every 5 steps of the training, we set the domainness variable $z$ as $[1,0,0,0]$, $[0,1,0,0]$, $[0,0,1,0]$, $[0,0,0,1]$ and uniformly distributed random variable. 
The qualitative results of the style generalization are shown in Fig~\ref{fig_example}. From the qualitative results, it is shown that our DLOW model can translate the photo image to corresponding artworks with different styles. When varying the values of domainness vector, we can also successfully produce new styles related to different painting styles, which demonstrates the good generalization ability of our model to unseen domains. Note, different from ~\cite{zhang2018unified, huang2017adain}, we do not need any reference image in the test phase, and the domainness vector can be changed instantly to generate different new styles of images. We provide more examples in Supplementary.

\textbf{Quantitative Results:} To verify the effectiveness of our model for style generalization, we conduct an user study on Amazon Mechanical Turk (AMT) to compare with the existing methods FadNet~\cite{lample2017fader} and MUNIT~\cite{huang2018munit}. Two cases are considered, style transfer to Van Gogh, and style generalization to mixed Van Gogh and Ukiyo-e. For FadNet, domain labels are treated as attributes. For MUNIT, we mix Van Gogh and Ukiyo-e as the target domain. The data for each trial is gathered from 10 participants and there are 100 trials in total for each case. For the first case, participants are shown the example Van Gogh style painting and are required to choose the image whose style is more similar to the example. For the second case, participants are shown the example Van Gogh and Ukiyo-e style painting and are required to choose the image with a style that is more like the mixed style of the two example paintings. The user preference is summarized in Table~\ref{tab:sg_res}, which shows that DLOW outperforms FadNet and MUNIT on both tasks. Qualitative comparison between different methods is provided in Supplementary due to the space limitation.

\begin{table}
\setlength{\tabcolsep}{3pt}
\small
    \centering
    \begin{tabular}{c|c|c}
    \hline
     & FadNet\cite{lample2017fader} / DLOW & MUNIT\cite{huang2018munit} / DLOW\\
    \hline
    Van Gogh &  $1.4\%$ / $98.6\%$ & $21.4\%$ / $78.6\%$\\
    Van Gogh + Ukiyo-e & $1.6\%$ / $98.4\%$ & $15.3\%$ / $84.7\%$ \\
    \hline
    \end{tabular}
    \caption{User preference for style transfer and generalization. It is shown that more users prefer our translated results on both of the style transfer and generalization tasks compared with the existing methods FadNet and MUNIT.}
    \label{tab:sg_res}
    \vspace{-10pt}
\end{table}
\section{Conclusion}
In this paper, we have presented the DLOW model to generate intermediate domains for bridging different domains. The model takes a domainness variable $z$ (or domainness vector $\z$) as the conditional input, and transfers images into the intermediate domain controlled by $z$ or $\z$. We demonstrate the benefits of our DLOW model in two scenarios. Firstly, for the cross-domain semantic segmentation task, our DLOW model can improve the performance of the pixel-level domain adaptation by taking the translated images in intermediate domains as training data. Secondly, our DLOW model also exhibits excellent style generalization ability for image translation and we are able to transfer images into a new style that is unseen in the training data. Extensive experiments on benchmark datasets have verified the effectiveness of our proposed model.

\vspace{5mm}
\noindent\textbf{Acknowledgments} The authors gratefully acknowledge the support by armasuisse.

\newpage
{\small
\bibliographystyle{ieee}
\bibliography{egbib}

\begin{thebibliography}{10}\itemsep=-1pt

\bibitem{almahairi2018augmented}
Amjad Almahairi, Sai Rajeswar, Alessandro Sordoni, Philip Bachman, and Aaron
  Courville.
\newblock Augmented {C}ycle{GAN}: Learning many-to-many mappings from unpaired
  data.
\newblock In {\em ICML}, 2018.

\bibitem{arjovsky2017wasserstein}
Martin Arjovsky, Soumith Chintala, and L{\'e}on Bottou.
\newblock Wasserstein gan.
\newblock {\em arXiv:1701.07875}, 2017.

\bibitem{baktashmotlagh2013unsupervised}
Mahsa Baktashmotlagh, Mehrtash~T Harandi, Brian~C Lovell, and Mathieu Salzmann.
\newblock Unsupervised domain adaptation by domain invariant projection.
\newblock In {\em ICCV}, 2013.

\bibitem{chen2018deeplab}
Liang-Chieh Chen, George Papandreou, Iasonas Kokkinos, Kevin Murphy, and Alan~L
  Yuille.
\newblock Deeplab: Semantic image segmentation with deep convolutional nets,
  atrous convolution, and fully connected crfs.
\newblock {\em IEEE transactions on pattern analysis and machine intelligence},
  40(4):834--848, 2018.

\bibitem{chen2018learning}
Yuhua Chen, Wen Li, Xiaoran Chen, and Luc Van~Gool.
\newblock Learning semantic segmentation from synthetic data: A geometrically
  guided input-output adaptation approach.
\newblock {\em arXiv:1812.05040}, 2018.

\bibitem{chen2018domain}
Yuhua Chen, Wen Li, Christos Sakaridis, Dengxin Dai, and Luc Van~Gool.
\newblock Domain adaptive faster r-cnn for object detection in the wild.
\newblock In {\em CVPR}, 2018.

\bibitem{chen2018road}
Yuhua Chen, Wen Li, and Luc Van~Gool.
\newblock Road: Reality oriented adaptation for semantic segmentation of urban
  scenes.
\newblock In {\em CVPR}, 2018.

\bibitem{StarGAN2018}
Yunjey Choi, Minje Choi, Munyoung Kim, Jung-Woo Ha, Sunghun Kim, and Jaegul
  Choo.
\newblock Star{GAN}: Unified generative adversarial networks for multi-domain
  image-to-image translation.
\newblock In {\em CVPR}, 2018.

\bibitem{Cordts2016Cityscapes}
Marius Cordts, Mohamed Omran, Sebastian Ramos, Timo Rehfeld, Markus Enzweiler,
  Rodrigo Benenson, Uwe Franke, Stefan Roth, and Bernt Schiele.
\newblock The cityscapes dataset for semantic urban scene understanding.
\newblock In {\em CVPR}, 2016.

\bibitem{cui2014flowing}
Zhen Cui, Wen Li, Dong Xu, Shiguang Shan, Xilin Chen, and Xuelong Li.
\newblock Flowing on riemannian manifold: Domain adaptation by shifting
  covariance.
\newblock {\em IEEE transactions on cybernetics}, 44(12):2264--2273, 2014.

\bibitem{dundar2018domain}
Aysegul Dundar, Ming-Yu Liu, Ting-Chun Wang, John Zedlewski, and Jan Kautz.
\newblock Domain stylization: A strong, simple baseline for synthetic to real
  image domain adaptation.
\newblock {\em arXiv:1807.09384}, 2018.

\bibitem{fernando2013unsupervised}
Basura Fernando, Amaury Habrard, Marc Sebban, and Tinne Tuytelaars.
\newblock Unsupervised visual domain adaptation using subspace alignment.
\newblock In {\em ICCV}, 2013.

\bibitem{ganin2015unsupervised}
Yaroslav Ganin and Victor~S. Lempitsky.
\newblock Unsupervised domain adaptation by backpropagation.
\newblock In {\em ICML}, 2015.

\bibitem{Geiger2012CVPR}
Andreas Geiger, Philip Lenz, and Raquel Urtasun.
\newblock Are we ready for autonomous driving? the kitti vision benchmark
  suite.
\newblock In {\em CVPR}, 2012.

\bibitem{ghifary2015domain}
Muhammad Ghifary, W Bastiaan~Kleijn, Mengjie Zhang, and David Balduzzi.
\newblock Domain generalization for object recognition with multi-task
  autoencoders.
\newblock In {\em ICCV}, 2015.

\bibitem{gong2012geodesic}
Boqing Gong, Yuan Shi, Fei Sha, and Kristen Grauman.
\newblock Geodesic flow kernel for unsupervised domain adaptation.
\newblock In {\em CVPR}, 2012.

\bibitem{goodfellow2014generative}
Ian Goodfellow, Jean Pouget-Abadie, Mehdi Mirza, Bing Xu, David Warde-Farley,
  Sherjil Ozair, Aaron Courville, and Yoshua Bengio.
\newblock Generative adversarial nets.
\newblock In {\em NIPS}, 2014.

\bibitem{gopalan2011domain}
Raghuraman Gopalan, Ruonan Li, and Rama Chellappa.
\newblock Domain adaptation for object recognition: An unsupervised approach.
\newblock In {\em ICCV}, 2011.

\bibitem{he2016deep}
Kaiming He, Xiangyu Zhang, Shaoqing Ren, and Jian Sun.
\newblock Deep residual learning for image recognition.
\newblock In {\em CVPR}, 2016.

\bibitem{he2017arbitrary}
Zhenliang He, Wangmeng Zuo, Meina Kan, Shiguang Shan, and Xilin Chen.
\newblock Arbitrary facial attribute editing: Only change what you want.
\newblock {\em arXiv:1711.10678}, 2017.

\bibitem{Hoffman_cycada2017}
Judy Hoffman, Eric Tzeng, Taesung Park, Jun-Yan Zhu, Phillip Isola, Kate
  Saenko, Alexei Efros, and Trevor Darrell.
\newblock {C}y{CADA}: Cycle-consistent adversarial domain adaptation.
\newblock In {\em ICML}, 2018.

\bibitem{hoffman2016fcns}
Judy Hoffman, Dequan Wang, Fisher Yu, and Trevor Darrell.
\newblock Fcns in the wild: Pixel-level adversarial and constraint-based
  adaptation.
\newblock {\em arXiv:1612.02649}, 2016.

\bibitem{hong2018conditional}
Weixiang Hong, Zhenzhen Wang, Ming Yang, and Junsong Yuan.
\newblock Conditional generative adversarial network for structured domain
  adaptation.
\newblock In {\em CVPR}, 2018.

\bibitem{zhang2018mixup}
Yann N. Dauphin David Lopez-Paz Hongyi~Zhang, Moustapha~Cisse.
\newblock mixup: Beyond empirical risk minimization.
\newblock In {\em ICLR}, 2018.

\bibitem{huang2018domain}
Haoshuo Huang, Qixing Huang, and Philipp Kr{\"a}henb{\"u}hl.
\newblock Domain transfer through deep activation matching.
\newblock In {\em ECCV}, 2018.

\bibitem{huang2017adain}
Xun Huang and Serge Belongie.
\newblock Arbitrary style transfer in real-time with adaptive instance
  normalization.
\newblock In {\em ICCV}, 2017.

\bibitem{huang2018munit}
Xun Huang, Ming-Yu Liu, Serge Belongie, and Jan Kautz.
\newblock Multimodal unsupervised image-to-image translation.
\newblock In {\em ECCV}, 2018.

\bibitem{pix2pix2017}
Phillip Isola, Jun-Yan Zhu, Tinghui Zhou, and Alexei~A Efros.
\newblock Image-to-image translation with conditional adversarial networks.
\newblock In {\em CVPR}, 2017.

\bibitem{jhuo2012robust}
I-Hong Jhuo, Dong Liu, DT Lee, and Shih-Fu Chang.
\newblock Robust visual domain adaptation with low-rank reconstruction.
\newblock In {\em CVPR}, 2012.

\bibitem{pmlr-v70-kim17a}
Taeksoo Kim, Moonsu Cha, Hyunsoo Kim, Jung~Kwon Lee, and Jiwon Kim.
\newblock Learning to discover cross-domain relations with generative
  adversarial networks.
\newblock In {\em ICML}, 2017.

\bibitem{kodirov2015unsupervised}
Elyor Kodirov, Tao Xiang, Zhenyong Fu, and Shaogang Gong.
\newblock Unsupervised domain adaptation for zero-shot learning.
\newblock In {\em ICCV}, 2015.

\bibitem{lample2017fader}
Guillaume Lample, Neil Zeghidour, Nicolas Usunier, Antoine Bordes, Ludovic
  DENOYER, et~al.
\newblock Fader networks: Manipulating images by sliding attributes.
\newblock In {\em NIPS}, 2017.

\bibitem{DRIT}
Hsin-Ying Lee, Hung-Yu Tseng, Jia-Bin Huang, Maneesh~Kumar Singh, and
  Ming-Hsuan Yang.
\newblock Diverse image-to-image translation via disentangled representations.
\newblock In {\em ECCV}, 2018.

\bibitem{Li2018Domain}
Haoliang Li, Sinno~Jialin Pan, Shiqi Wang, and Alex~C Kot.
\newblock Domain generalization with adversarial feature learning.
\newblock In {\em CVPR}, 2018.

\bibitem{li2018domain_gene}
Wen Li, Zheng Xu, Dong Xu, Dengxin Dai, and Luc Van~Gool.
\newblock Domain generalization and adaptation using low rank exemplar svms.
\newblock {\em IEEE transactions on pattern analysis and machine intelligence},
  40(5):1114--1127, 2018.

\bibitem{li2018deep}
Ya Li, Xinmei Tian, Mingming Gong, Yajing Liu, Tongliang Liu, Kun Zhang, and
  Dacheng Tao.
\newblock Deep domain generalization via conditional invariant adversarial
  networks.
\newblock In {\em ECCV}, 2018.

\bibitem{lin2018conditional}
Jianxin Lin, Yingce Xia, Tao Qin, Zhibo Chen, and Tie-Yan Liu.
\newblock Conditional image-to-image translation.
\newblock In {\em CVPR}, 2018.

\bibitem{NIPS2017_6672}
Ming-Yu Liu, Thomas Breuel, and Jan Kautz.
\newblock Unsupervised image-to-image translation networks.
\newblock In {\em NIPS}, 2017.

\bibitem{lu2017conditional}
Yongyi Lu, Yu-Wing Tai, and Chi-Keung Tang.
\newblock Conditional cyclegan for attribute guided face image generation.
\newblock {\em arXiv:1705.09966}, 2017.

\bibitem{luo2018taking}
Yawei Luo, Liang Zheng, Tao Guan, Junqing Yu, and Yi Yang.
\newblock Taking a closer look at domain shift: Category-level adversaries for
  semantics consistent domain adaptation.
\newblock {\em arXiv:1809.09478}, 2018.

\bibitem{motiian2017unified}
Saeid Motiian, Marco Piccirilli, Donald~A Adjeroh, and Gianfranco Doretto.
\newblock Unified deep supervised domain adaptation and generalization.
\newblock In {\em ICCV}, 2017.

\bibitem{pmlr-v28-muandet13}
Krikamol Muandet, David Balduzzi, and Bernhard Schölkopf.
\newblock Domain generalization via invariant feature representation.
\newblock In {\em ICML}, 2013.

\bibitem{murez2017image}
Zak Murez, Soheil Kolouri, David Kriegman, Ravi Ramamoorthi, and Kyungnam Kim.
\newblock Image to image translation for domain adaptation.
\newblock {\em arXiv:1712.00479}, 2017.

\bibitem{niu2015multi}
Li Niu, Wen Li, and Dong Xu.
\newblock Multi-view domain generalization for visual recognition.
\newblock In {\em ICCV}, 2015.

\bibitem{niu2015visual}
Li Niu, Wen Li, and Dong Xu.
\newblock Visual recognition by learning from web data: A weakly supervised
  domain generalization approach.
\newblock In {\em CVPR}, 2015.

\bibitem{pan2018IBN-Net}
Xingang Pan, Ping Luo, Jianping Shi, and Xiaoou Tang.
\newblock Two at once: Enhancing learning and generalization capacities via
  ibn-net.
\newblock In {\em ECCV}, 2018.

\bibitem{peng2018visda}
Xingchao Peng, Ben Usman, Neela Kaushik, Dequan Wang, Judy Hoffman, Kate
  Saenko, Xavier Roynard, Jean-Emmanuel Deschaud, Francois Goulette, Tyler~L
  Hayes, et~al.
\newblock Visda: A synthetic-to-real benchmark for visual domain adaptation.
\newblock In {\em CVPR Workshops}, 2018.

\bibitem{Richter_2016_ECCV}
Stephan~R. Richter, Vibhav Vineet, Stefan Roth, and Vladlen Koltun.
\newblock Playing for data: {G}round truth from computer games.
\newblock In {\em ECCV}, 2016.

\bibitem{Ros_2016_CVPR}
German Ros, Laura Sellart, Joanna Materzynska, David Vazquez, and Antonio~M.
  Lopez.
\newblock The synthia dataset: A large collection of synthetic images for
  semantic segmentation of urban scenes.
\newblock In {\em CVPR}, 2016.

\bibitem{saito2017maximum}
Kuniaki Saito, Kohei Watanabe, Yoshitaka Ushiku, and Tatsuya Harada.
\newblock Maximum classifier discrepancy for unsupervised domain adaptation.
\newblock {\em arXiv:1712.02560}, 2017.

\bibitem{saleh2018effective}
Fatemeh~Sadat Saleh, Mohammad~Sadegh Aliakbarian, Mathieu Salzmann, Lars
  Petersson, and Jose~M Alvarez.
\newblock Effective use of synthetic data for urban scene semantic
  segmentation.
\newblock In {\em ECCV}, 2018.

\bibitem{sankaranarayanan2017unsupervised}
Swami Sankaranarayanan, Yogesh Balaji, Arpit Jain, Ser~Nam Lim, and Rama
  Chellappa.
\newblock Unsupervised domain adaptation for semantic segmentation with gans.
\newblock {\em arXiv preprint arXiv:1711.06969}, 2017.

\bibitem{sankaranarayanan2018learning}
Swami Sankaranarayanan, Yogesh Balaji, Arpit Jain, Ser~Nam Lim, and Rama
  Chellappa.
\newblock Learning from synthetic data: Addressing domain shift for semantic
  segmentation.
\newblock In {\em CVPR}, 2018.

\bibitem{Tsai_adaptseg_2018}
Y.-H. Tsai, W.-C. Hung, S. Schulter, K. Sohn, M.-H. Yang, and M. Chandraker.
\newblock Learning to adapt structured output space for semantic segmentation.
\newblock In {\em CVPR}, 2018.

\bibitem{tzeng2017adversarial}
Eric Tzeng, Judy Hoffman, Kate Saenko, and Trevor Darrell.
\newblock Adversarial discriminative domain adaptation.
\newblock In {\em CVPR}, 2017.

\bibitem{wang2018pix2pixHD}
Ting-Chun Wang, Ming-Yu Liu, Jun-Yan Zhu, Andrew Tao, Jan Kautz, and Bryan
  Catanzaro.
\newblock High-resolution image synthesis and semantic manipulation with
  conditional gans.
\newblock In {\em CVPR}, 2018.

\bibitem{wulfmeier2018incremental}
Markus Wulfmeier, Alex Bewley, and Ingmar Posner.
\newblock Incremental adversarial domain adaptation for continually changing
  environments.
\newblock In {\em ICRA}, 2018.

\bibitem{yi2017dualgan}
Zili Yi, Hao~(Richard) Zhang, Ping Tan, and Minglun Gong.
\newblock Dualgan: Unsupervised dual learning for image-to-image translation.
\newblock In {\em ICCV}, 2017.

\bibitem{yu2018bdd100k}
Fisher Yu, Wenqi Xian, Yingying Chen, Fangchen Liu, Mike Liao, Vashisht
  Madhavan, and Trevor Darrell.
\newblock Bdd100k: A diverse driving video database with scalable annotation
  tooling.
\newblock {\em arXiv:1805.04687}, 2018.

\bibitem{Zendel_2018_ECCV}
Oliver Zendel, Katrin Honauer, Markus Murschitz, Daniel Steininger, and Gustavo
  Fernandez~Dominguez.
\newblock Wilddash - creating hazard-aware benchmarks.
\newblock In {\em ECCV}, 2018.

\bibitem{zhang2018collaborative}
Weichen Zhang, Wanli Ouyang, Wen Li, and Dong Xu.
\newblock Collaborative and adversarial network for unsupervised domain
  adaptation.
\newblock In {\em CVPR}, 2018.

\bibitem{zhang2018fully}
Yiheng Zhang, Zhaofan Qiu, Ting Yao, Dong Liu, and Tao Mei.
\newblock Fully convolutional adaptation networks for semantic segmentation.
\newblock In {\em CVPR}, 2018.

\bibitem{zhang2018unified}
Yexun Zhang, Ya Zhang, and Wenbin Cai.
\newblock A unified framework for generalizable style transfer: Style and
  content separation.
\newblock {\em arXiv:1806.05173}, 2018.

\bibitem{zhu2017unpaired}
Jun-Yan Zhu, Taesung Park, Phillip Isola, and Alexei~A Efros.
\newblock Unpaired image-to-image translation using cycle-consistent
  adversarial networks.
\newblock In {\em ICCV}, 2017.

\bibitem{zhu2017toward}
Jun-Yan Zhu, Richard Zhang, Deepak Pathak, Trevor Darrell, Alexei~A Efros,
  Oliver Wang, and Eli Shechtman.
\newblock Toward multimodal image-to-image translation.
\newblock In {\em NIPS}, 2017.

\bibitem{zhu2018penalizing}
Xinge Zhu, Hui Zhou, Ceyuan Yang, Jianping Shi, and Dahua Lin.
\newblock Penalizing top performers: Conservative loss for semantic
  segmentation adaptation.
\newblock {\em arXiv:1809.00903}, 2018.

\bibitem{zou2018unsupervised}
Yang Zou, Zhiding Yu, BVK~Vijaya Kumar, and Jinsong Wang.
\newblock Unsupervised domain adaptation for semantic segmentation via
  class-balanced self-training.
\newblock In {\em ECCV}, 2018.

\end{thebibliography}
}

\clearpage{}
\clearpage{
\section{Supplementary}
In this Supplementary, we provide additional information for,
\begin{itemize}
    \item enlarged version of DLOW translated images for GTA5 to Cityscapes,
    \item the adaptation and generalization performance of our DLOW model on SYNTHIA to Cityscapes,
    \item the detailed network structure of our DLOW model for style generalization with four target domains,
    \item more examples for style generalization,
    \item the qualitative comparison of different methods on style transfer and style generalization.
\end{itemize}

\subsection{Comparison of DLOW Translated Images with Brightness Adjusted Images}
In Section 4.1 of the main paper, we show the examples of intermediate domain images between the source domain GTA5 and the target domain Cityscapes. The main change in those images at the first glance might be the image brightness. Here we provide an enlarged version of intermediate images to show that not only the brightness but also the subtle texture are adjusted to mimic the Cityscapes style. For comparison, we adjust the brightness of the translated image with $z=0$ to match it with the brightness of the corresponding translated image with $z=1$. The enlarged translated image with $z=1$ and the corresponding the brightness adjusted image($z=0$) are shown in Fig~\ref{interimage_comparison:fig}, from which we observe that the brightness adjusted image still exhibits obvious features of the game style such as the high contrast textures of the road and the red curb, while our DLOW translated image well mimics the texture of Cityscapes style.

\subsection{Additional Results for Domain Adaptation and Generalization}
In Section 4.1 of the main paper, we show the adaptation and the generalization performance of the DLOW model on the GTA5 to Cityscapes dataset. In this Supplementary, we further present the experimental results of our DLOW model on the SYNTHIA to Cityscapes dataset. The SYNTHIA dataset \cite{Ros_2016_CVPR} is used as the source domain while the Cityscapes dataset \cite{Cordts2016Cityscapes} is used as the target domain. Similar to the experiment on GTA5, we also evaluate the generalization ability of learnt segmentation models to unseen domains on the KITTI~\cite{Geiger2012CVPR}, WildDash~\cite{Zendel_2018_ECCV} and BDD100K~\cite{yu2018bdd100k} datasets.

\textbf{SYNTHIA-RAND-CITYSCAPES} is a dataset comprising 9400 photo-realistic images rendered from a virtual city and the semantic labels of the images are precise and compatible with Cityscapes test set.

The same training parameters and scheme as GTA5 are applied to SYNTHIA dataset, while the only difference lies in that we resize the training images to $1280\times 760$ for the segmentation network.

Similar to GTA5, our DLOW model based on SYNTHIA dataset also exhibits excellent performance for the domain adaptation and the domain generalization. Following \cite{Tsai_adaptseg_2018}, the segmentation performance based on SYNTHIA dataset is tested on the Cityscapes validation dataset with 13 classes. As shown in Table \ref{table_pixel_synthia}, all pixel-level adaptation methods outperform the ``NonAdapt" baseline, which verifies the effectiveness of the image translation for cross-domain segmentation. In particular, our ``DLOW($z=1$)" model achieves $41.6\%$, gaining $3\%$ improvment compared to the `NonAdapt" baseline. After using the intermediate domain images, the adaptation performance can be further improved from $41.6\%$ to $42.8\%$. The Table \ref{table_generalization_synthia} also reports the result of our DLOW model adaptation performance combining with the AdaptSegNet method and the domain generalization performance for the unseen domains. The Original$^*$ in Table \ref{table_generalization_synthia} denotes our retrained multi-level AdaptSegNet model in \cite{Tsai_adaptseg_2018}. Compared with the retraining AdaptSegNet model, our DLOW model could improve the adaptation performance from $45.7\%$ to $47.1\%$. The domain generalization results show that the intermediate domain images could improve the generalization ability of the adapted model.

\begin{table}
\setlength{\tabcolsep}{3pt}
\centering
 \caption{\label{table_generalization_synthia} Comparison of the performance of AdaptSegNet~\cite{Tsai_adaptseg_2018} when using original source images and intermediate domain images translated with our DLOW model for semantic segmention under domain adaptation (1st column) and domain generalization (2nd to 4th columns) scenarios. The Original$^*$ denotes our retrained multi-level AdaptSegNet model. The Original model is provided by the author of AdaptSegNet. The results are reported on mIoU over 13 categories. The best result is denoted in bold.}
 \begin{tabular}{c|cccc}
  \hline
  &Cityscapes&KITTI&WildDash&BDD100K\\
  \hline
  Original~\cite{Tsai_adaptseg_2018}&46.7&33.3&20.6&30.8\\
  Original$^*$~\cite{Tsai_adaptseg_2018}&45.7&\textbf{34.4}&20.0&30.8\\
  DLOW&\textbf{47.1}&\textbf{34.4}&\textbf{24.4}&\textbf{35.3}\\
  \hline
 \end{tabular}
\end{table}
\begin{table*}[h]
\centering
 \caption{\label{table_pixel_synthia} Results of semantic segmentation on the CityScapes dataset based on DeepLab-v2 model with ResNet-101 backbone using the images translated with different models. The results are reported on mIoU over 13 categories. The best result is denoted in bold.}
\begin{tabular}{c|ccccccccccccc|c}
  \hline
  \multicolumn{15}{c}{SYNTHIA $\rightarrow$ Cityscapes}\\
  \hline
  Method&\rotatebox{90}{road}&\rotatebox{90}{sidewalk}&\rotatebox{90}{building}&\rotatebox{90}{traffic light}&\rotatebox{90}{traffic sign}&\rotatebox{90}{vegetation}&\rotatebox{90}{sky}&\rotatebox{90}{person}&\rotatebox{90}{rider}&\rotatebox{90}{car}&\rotatebox{90}{bus}&\rotatebox{90}{motorbike}&\rotatebox{90}{bicycle}&mIoU\\
  \hline
  NonAdapt\cite{Tsai_adaptseg_2018}&55.6&23.8&74.6&6.1&12.1&74.8&79.0&\textbf{55.3}&19.1&39.6&23.3&13.7&25.0&38.6\\
CycleGAN\cite{Hoffman_cycada2017}&69.4&\textbf{28.3}&73.8&12.7&15.2&74.0&78.9&46.2&18.0&62.2&\textbf{27.6}&14.2&27.2&42.1\\
  \hline
  DLOW($z=1$)&\textbf{71.0}&26.8&74.0&\textbf{13.9}&\textbf{17.5}&75.6&79.9&43.5&17.0&63.5&16.7&\textbf{14.5}&27.4&41.6\\
  DLOW&65.3&22.4&\textbf{75.5}&9.1&13.2&\textbf{76.1}&\textbf{80.4}&52.0&\textbf{21.1}&\textbf{70.5}&26.3&10.7&\textbf{33.5}&\textbf{42.8}\\
  \hline
 \end{tabular}
\end{table*}

\begin{figure*}
  \centering
  \includegraphics[height=0.25\paperheight]{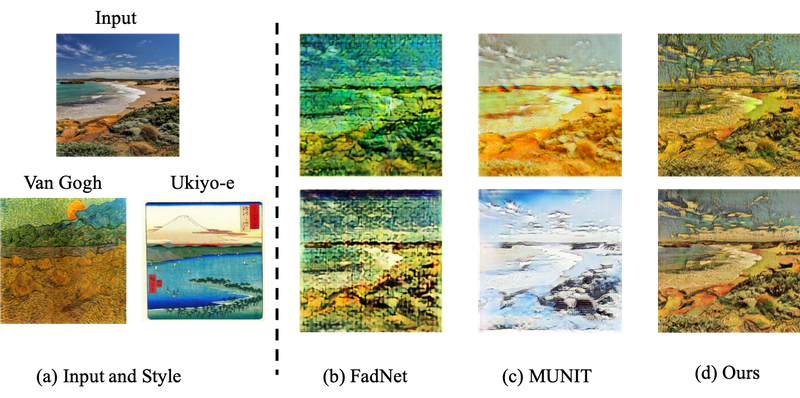}
  \caption{Comparison of our model with existing methods on style transfer and style generalization. The left part (a) shows the given input photo image and the example images of the target style. The translated results with different methods FadNet~\cite{lample2017fader}, MUNIT~\cite{huang2018munit} and our DLOW are shown in right part (b), (c) and (d). The first row of the right part is the Van Gogh style transfer result while the second row is the style generalization result aiming at mixing the Van Gogh and Ukiyo-e style.}
  \label{fig_style_comparison}
\end{figure*}

\subsection{Network Structure for Style Generalization }
In Section 3.6 of the main paper, we introduce that our DLOW model can be adapted for style generalization when there are multiple target domains available. We present the details in this section. The network structure of our DLOW model for style generalization is shown in Fig~\ref{style_gen:fig}, where we have four target domains, each of which represents an image style. For the direction of $\cS\rightarrow\cT$, shown in Fig~\ref{style_gen_supp:fig1}, the style generalization model consists of two modules, the adversarial module and the image reconstruction module. For each target domain $\cT_{i}$, there is one corresponding discriminator $D_{T_{i}}$ measuring the distribution distance between the intermediate domain $\cM^{(z)}$ and the target domain $\cT_{i}$. Accordingly, the domainness variable $z$ is expanded as a $4$-dim vector $\z = [z_1, \ldots, z_4]'$. For the other direction $\cT\rightarrow\cS$, shown in Fig~\ref{style_gen_supp:fig2}, the adversarial module is similar to that of the direction $\cS\rightarrow\cT$. However, the image reconstruction module is slightly different, since the image reconstruction loss should be weighted by the domainness vector $\z$.

\subsection{Additional Results for Style Generalization}
We provide an example for style generalization in Fig 6 of the main paper. Here we provide more experimental results in Fig~\ref{fig_arbit_append:fig} and Fig~\ref{fig_arbit_append2:fig}. The images with red bounding boxes are translated images in four target domains, \ie,  Monet, Van Gogh, Cezanne, and Ukiyo-e. Those can be considered as the ``seen" styles. Our model gives similar translation results to CycleGAN model for each target domain. But the difference is that we only need one unified model for the four target domains while the CycleGAN should train four models. Moreover, the images with green bounding boxes are the mixed style images of their neighboring target styles and the image in the center is the mixed style image of all the four target styles, which are new styles that are never seen in the training data. We can observe that our DLOW model could generalize well across different styles, which proves the good domain generalization ability of our model. 

\subsection{Qualitative Comparison for Style Transfer and Style Generalization}
In Section 4.2 of the main paper, we show the quantitative comparison results of our DLOW model with the FadNet~\cite{lample2017fader} and the MUNIT~\cite{huang2018munit} on the style transfer and the style generalization task. In this Supplementary, we further provide the qualitative result comparison in Fig~\ref{fig_style_comparison}. It can be observed that the FadNet fails to translate the photo to painting while the MUNIT and our DLOW model both could get reasonable results. For the Van Gogh style transfer result shown in Fig~\ref{fig_style_comparison}, our DLOW model could not only learn the typical color of the painting but also the details such as the brushwork and lines while the MUNIT only learns the main colors. For the Van Gogh and Ukiyo-e style generalization results shown in Fig~\ref{fig_style_comparison}, our DLOW model could combine the color and the stroke of the two styles while the MUNIT just fully changes the main colors from one style to another. The qualitative comparison result also demonstrates that our DLOW model performs better on both of the style transfer and generalization task compared with the FadNet and the MUNIT.
\begin{figure*}[!ht]
\centering
\begin{subfigure}[c]{\textwidth}
  \centering
  \includegraphics[width=\linewidth]{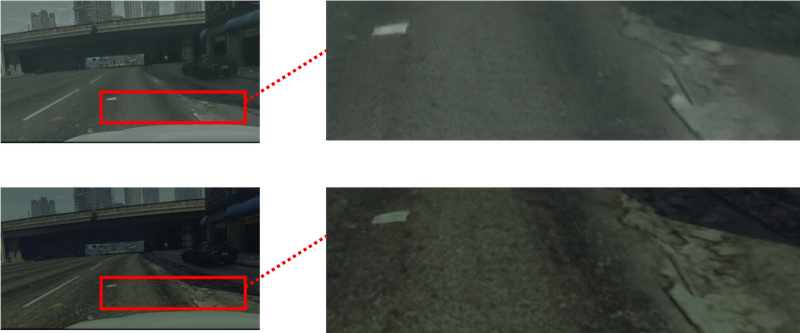}
  \label{style_gen:fig1}
\end{subfigure}
\begin{subfigure}[c]{\textwidth}
  \centering
  \includegraphics[width=\linewidth]{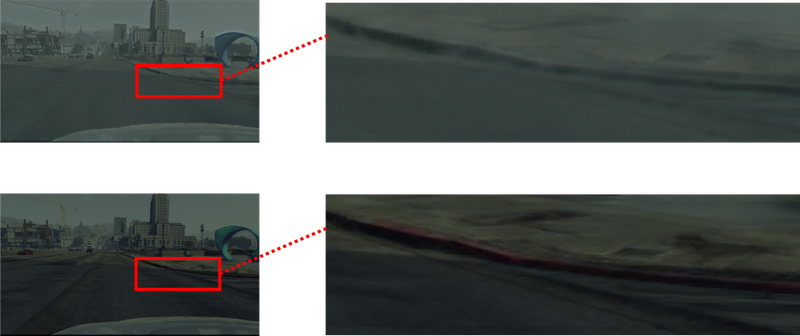}
  \label{style_gen:fig2}
\end{subfigure}
\caption{Examples of comparison between the DLOW translated image and the brightness adjusted image. We adjust the brightness of the DLOW translated source image($z=0$) to make its brightness match the corresponding DLOW translated target image($z=1$). The lower one in each group is the brightness adjusted image while the upper one is the DLOW translated target image($z=1$). Part of the image is enlarged and shown in the right to prove that our DLOW translation not only change the brightness but also change the details such as the texture of the road and the style of the curb to mimic the feature of the Cityscapes image.}
\label{interimage_comparison:fig}
\end{figure*}

\begin{figure*}[!ht]
\centering
\begin{subfigure}[c]{\textwidth}
  \centering
  \includegraphics[width=0.8\linewidth]{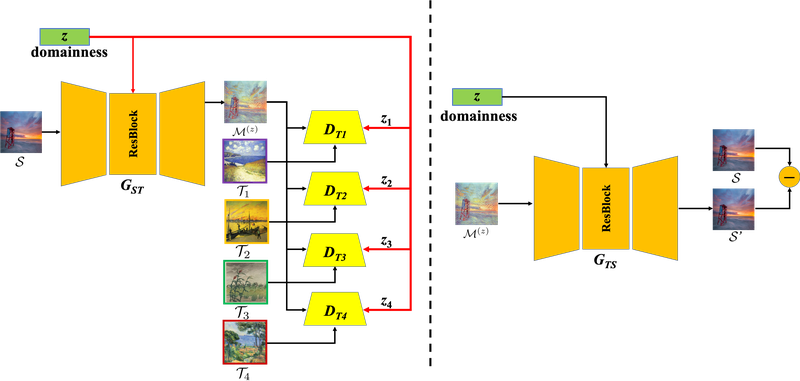}
  \caption{}
  \label{style_gen_supp:fig1}
\end{subfigure}
\begin{subfigure}[c]{\textwidth}
  \centering
  \includegraphics[width=\linewidth]{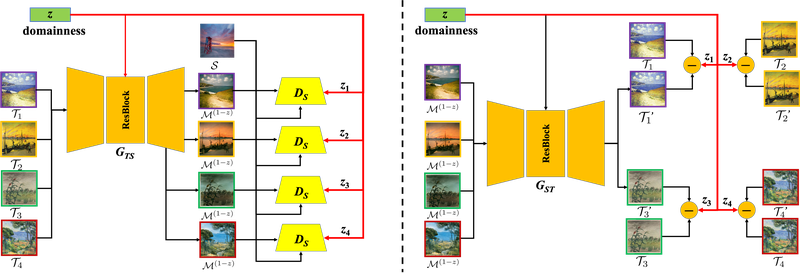}
  \caption{}
  \label{style_gen_supp:fig2}
\end{subfigure}
\caption{Network structure of DLOW model for style generalization with four target domains: (a) direction from $\cS\rightarrow\cT$; (b) direction from $\cT\rightarrow\cS$.}
\label{style_gen:fig}
\end{figure*}

\begin{figure*}
\centering
\begin{subfigure}[c]{\textwidth}
  \centering
\includegraphics[height=0.39\paperheight]{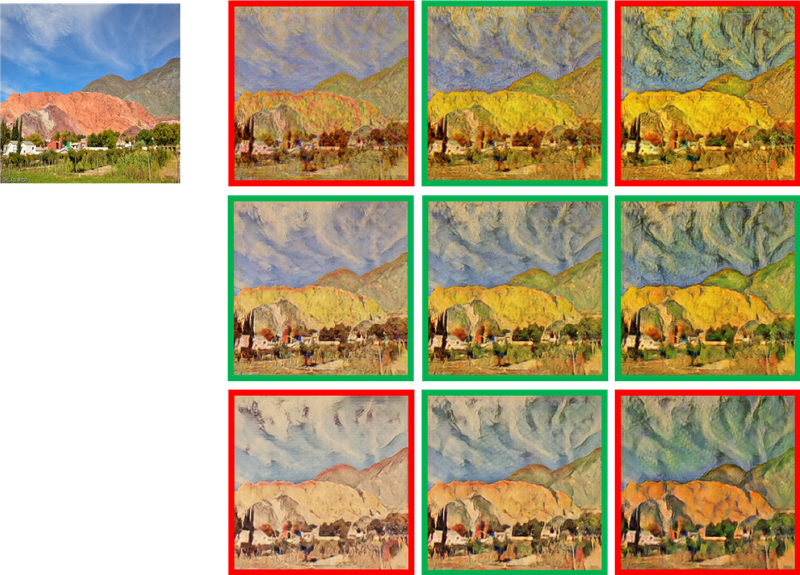}
\end{subfigure}
\begin{subfigure}[c]{\textwidth}
  \centering
\includegraphics[height=0.39\paperheight]{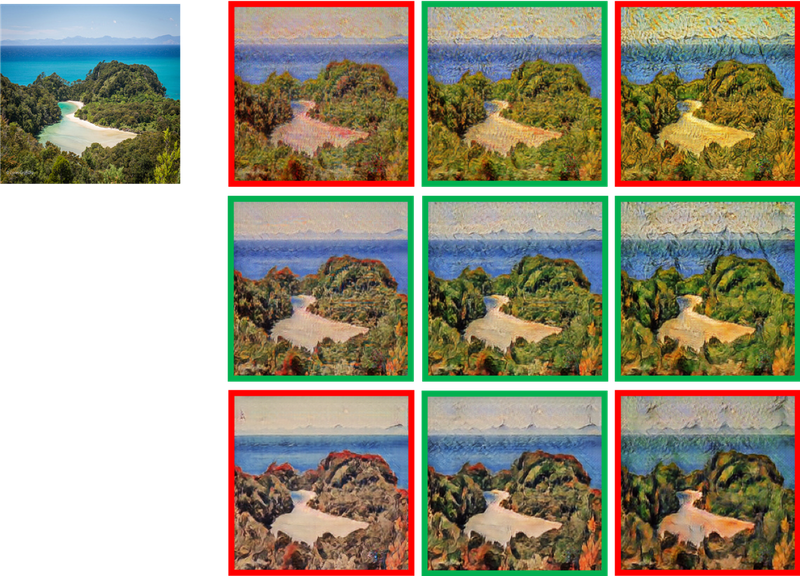}
\end{subfigure}
\vspace{-10pt}
\caption{Examples of style generalization I. Results with red rectangles at four corners are images translated into the four target domains, and those with green rectangles in between are images translated into intermediate domains. The results show that our DLOW model generalizes well across styles, and produces new images styles smoothly.}
\label{fig_arbit_append:fig}
\end{figure*}
\begin{figure*}
\centering
\begin{subfigure}[c]{\textwidth}
  \centering
\includegraphics[height=0.39\paperheight]{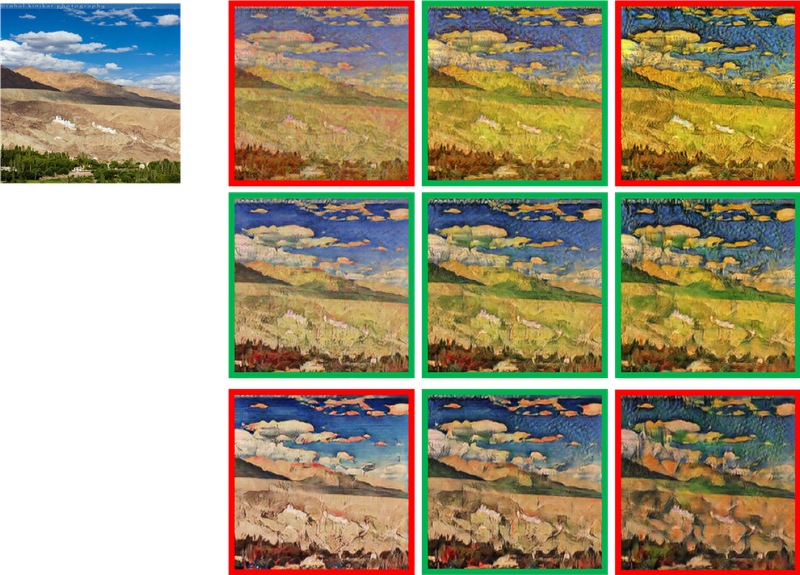}
\end{subfigure}
\begin{subfigure}[c]{\textwidth}
  \centering
\includegraphics[height=0.39\paperheight]{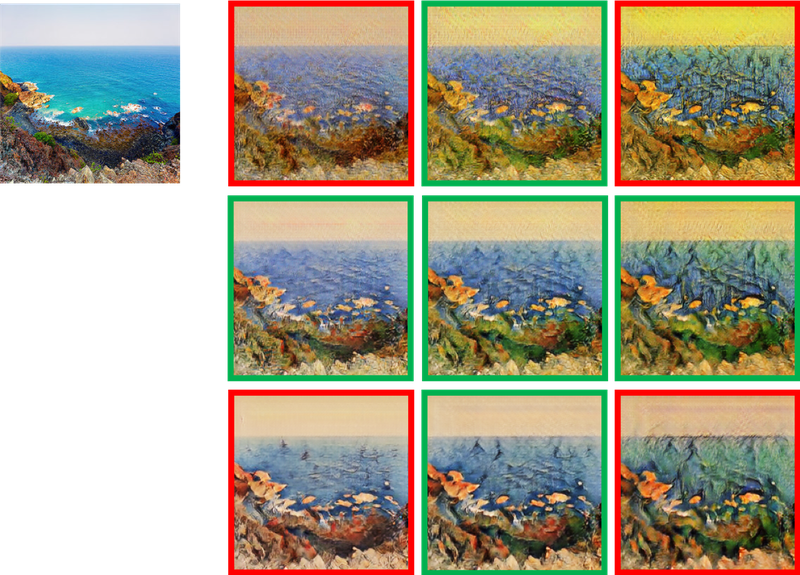}
\end{subfigure}
\vspace{-10pt}
\caption{Examples of style generalization II. Results with red rectangles at four corners are images translated into the four target domains, and those with green rectangles in between are images translated into intermediate domains. The results show that our DLOW model generalizes well across styles, and produces new images styles smoothly.}
\label{fig_arbit_append2:fig}
\end{figure*}
}
\end{document}